\documentclass{article}
\usepackage{ijcai22}

\usepackage{times}
\usepackage{soul}
\usepackage{url}
\usepackage[hidelinks]{hyperref}
\usepackage[utf8]{inputenc}
\usepackage[small]{caption}
\usepackage{graphicx}
\usepackage{amsmath,amssymb,amsfonts}
\usepackage{amsthm}
\usepackage{booktabs}
\usepackage{algorithm}
\usepackage{algorithmic}
\usepackage{textcomp}
\usepackage{xcolor}
\usepackage{svg}

\usepackage{todonotes}
\usepackage{bm}
\usepackage{multirow}

\usepackage{pgfplots} 
\usepackage{definitions, defs}
\usepgfplotslibrary{groupplots}
\pgfplotsset{compat=newest}
\usepackage{tikzscale}
\usepackage{xspace}
\usepackage{color, colortbl}

\newcommand{\agg}{A}

\newcommand{\fP}{\hat{P}}
\newcommand{\var}{\hat{\sigma}}
\newcommand{\mean}{\hat{\mu}}
\newcommand{\win}[1]{{w_{#1}}}

\newcommand{\trnlen}{l_{\mathrm{trn}}}
\newcommand{\vallen}{l_{\mathrm{val}}}
\newcommand{\testlen}{l_{\mathrm{test}}}

\newcommand{\Dtrn}{\mathcal{D}_{\mathrm{trn}}}
\newcommand{\Dval}{\mathcal{D}_{\mathrm{val}}}
\newcommand{\Dtest}{\mathcal{D}_{\mathrm{test}}}

\title{IJCAI--22 Formatting Instructions}

\author{
Prathamesh Deshpande\footnote{Contact author}
\and
Sunita Sarawagi
\affiliations
Department of Computer Science and Engineering, IIT Bombay
\emails
pratham@cse.iitb.ac.in,
sunita@iitb.ac.in
}

\begin{document}


\title{Coherent Probabilistic Aggregate Queries on Long-horizon Forecasts}

\maketitle


\begin{abstract}

Long range forecasts are the starting point of many decision support systems that need  to draw inference from high-level aggregate patterns on forecasted values. State of the art time-series forecasting methods are either subject to concept drift on long-horizon forecasts, or fail to accurately predict coherent and accurate high-level aggregates.

In this work, we present a novel probabilistic forecasting method that produces forecasts that are coherent in terms of base level and predicted aggregate statistics. We achieve the coherency between predicted base-level and aggregate statistics using a novel inference method based on KL-divergence that can be solved efficiently in closed form. We  show that our method improves forecast performance across both base level and unseen aggregates post inference on real datasets ranging three diverse domains. 
(\href{https://github.com/pratham16cse/AggForecaster}{Project URL})
%
\end{abstract}


\section{Introduction}


Long-term forecasting  is critical for decision support in several application domains including finance, governance, and climate studies.  Interactive decision making tools need accurate probability distributions not just over the long-term base forecasts but also on dynamically chosen aggregates of the forecasts.   An analyst may be interested in a variety of aggregates such as average, sum, or trend of values for various windows of granularity. For example, the base-level forecasts may be retail price of a commodity at a daily-level for the next two quarters, but the end user may also inspect average at weekly or monthly granularity, or change in price (inflation) from week to week. For the aggregates too, the user is interested in the associated uncertainty represented as a distribution that is coherent with the base distribution.    



A time-series in our model is characterized by a sequence  $(\vx_1, y_1)\ldots (\vx_n,y_n)$ where $\vx_t \in R^d$ denotes a vector of input features and  $y_t$ is a real-value at a time $t$. We are given a prefix or history from such a series up to a present time $T$ which we denote as $H_T$. Our goal is to predict future values $y_{T+1}\ldots y_{T+R}$ given input features $\vx=\vx_{T+1}\ldots \vx_{T+R}$ and the history $H_T$.  Unlike in conventional time-series forecasting where the interest is short-term (say $R$ between 1 and 10), in long-term forecasting we focus on large values of $R$ (say between 100 and 1000). We wish to capture the uncertainty of the prediction by outputting a distribution over the predicted values.  We use $\fP(y_{T+1}\ldots y_{T+R}|H_T, \vx)$ to denote the predicted distribution for future time horizon given history $H_T$. 

Our goal is not just to provide accurate probabilistic forecasts for the the base-level values but also for aggregates of these values. For example, the base-level forecasts may be at the daily-level and we may wish to view monthly sales. In theory, we could marginalize the base-level forecast distribution  $\fP(y_{T+1}\ldots y_{T+R}|H_T, \vx)$ to compute coherent distribution of any dynamic aggregate. In practice, we need to ensure that the marginalization can be efficiently computed. Furthermore, the accuracy of the predicted aggregate should be no worse than the accuracy of an independently trained model to directly predict the distribution of the aggregate.  For example, a practitioner could train separate models for predicting monthly sales and daily sales for greater accuracy at the monthly-level.  However, training separate models for each aggregate has two limitations: (1) independently predicted aggregate distributions may not be coherent with each other or with the base-level forecasts, and (2) the set of aggregate functions may be decided dynamically during data analysis where training a separate model is impractical.  Thus, we seek solutions that explicitly marginalize a fixed distribution  to ensure coherence of distributions of aggregate quantities.

Existing forecast models 
can be categorized as either auto-regressive (AR) ~\cite{FlunkertSG17,Mukherjee2018} or non-auto regressive (NAR).  We will show in this paper that both these models are inadequate for our goal of efficiently and dynamically defining the distribution of any aggregate quantity.
In AR models the output distribution captures the full joint dependency among all predicted values by defining that the value at $t$ depends on all values before it.   During training  we condition on true previous values but during inference we need to perform forward sampling where each predicted value is conditioned on sampled prior values. In long-term forecasting, such forward sampling suffers from drift caused by cascading errors \cite{deshpande2021DualTPP}.  
A second major limitation of the AR model is that the joint distribution over the forecast variables is expressed via a black-box neural network. Marginalizing the distribution to obtain distribution over dynamically chosen aggregates will involve repeated sampling steps, which may incur huge latency during interactive analysis.

This has led to greater interest in the non-auto-regressive (NAR) models which  independently predict each variable in the forecast horizon conditioned on the history. 
NAR models have been shown to work better in practice both for medium-term and long-term forecasting~~\cite{wen2017multi,Deshpande2019,zhou2020informer}. Another major benefit of the NAR models is that during inference all values can be predicted in parallel, unlike in the AR method that entails sequential sampling.
However, a limitation of the non-auto-regressive methods is that the dependency among the output variables is not modeled.   The distribution over aggregates computed from independent base predictions may be inaccurate.   For example, if an analyst marginalizes the data to inspect the distribution of the difference between adjacent values, the variance from the independent model might be over-estimated. Further, 
in the raw form a time-series often contains noise that hinders a forecasting model from capturing top-level patterns in the series over long prediction horizons. 
In applications that involve analysis over aggregates of forecast values, these might give rise to non-smooth or inconsistent aggregates. 

In this paper we propose a novel method of preserving higher-level patterns in long-range forecasts of non-auto-regressive forecast models while allowing accurate and efficient computation of the distribution of dynamically chosen aggregates.  Our key idea is to train the model to independently output various high-order statistics over the forecast values, along with the base-level forecasts.   Aggregated values provide a noise-removed signal of how the series evolves in time.  We use such signals to guide inference to capture top-level trends in the data. We formulate that as a task of obtaining a revised consensus distribution that minimizes the KL-distance with each predicted aggregate and base forecasts.   The consensus distribution is represented as a Gaussian distribution that can be easily marginalized. The parameters of the consensus can be obtained via an efficient numerical optimization problem.  Unlike existing methods~\cite{han21a}, our consensus distribution not only provides coherence and accuracy for aggregate quantities, but also improves base-level forecasts.


\subsection{Contributions}
\begin{itemize}
    \item We propose a probabilistic forecasting method that is more accurate than state of the art for long-term forecasting.
    \item Our method outputs an easy to marginalize joint distribution for computing accurate and coherent aggregate distributions over dynamically defined functions on forecast values.  
    \item We propose an efficient inference algorithm for computing the consensus distribution.
    \item Empirical evaluation on benchmark datasets show that our method improves both base-level and aggregate forecasts compared to several existing baselines including a recent hierarchical consensus model. 
\end{itemize}




\section{Our Model}

Our method seeks to retain the advantage of parallel training and inference that modern transformer-based NAR models offer while providing accurate distributions of forecasts at the base-level and dynamically chosen aggregates over them. 
We train multiple NAR models for forecasting different types of aggregate functions at  different levels of granularity.  During inference we obtain a consensus distribution, as a joint Gaussian distribution that can be marginalized in closed form for computing distributions over linear aggregates defined dynamically during analysis. The consensus distribution is obtained by minimizing the KL distance with the various base and aggregate-level forecasts by an efficient inference algorithm.  We present the details of each of these steps next.

\newcommand{\cA}{\mathcal{A}}
\subsection{Aggregate Functions}
\label{subsec:aggregate_functions}
We choose a set $\cA$ of aggregate functions depending on  likely use in downstream analysis. This set does not need to include all possible aggregate functions to be used during analysis, since our method generalizes well to new aggregates even when trained with two aggregate functions. Each aggregate function $\agg_i$ is characterized with a window size $K_i$
and a fixed real vector $\va^i\in \real^{K_i}$ that denotes the weight of aggregated values. For each $A_i$,
we create an aggregated series by aggregating on disjoint windows of size $K_i$ in the original series. 
The $j$-th value in the $i$th aggregated series is the aggregation of the values $y_{(j-1)K_i+1}\ldots y_{jK_i}$ and denoted $z^i_j$. We will use the notation $\win{i,j}$ to refer to the window of indices $[(j-1)K_i,\ldots,{j K_i}]$ that the $j$-th value of the $i$th function aggregates. 
%
We denote $i$-th aggregate series as $z^i_1, \ldots, z^i_{T_i}$ where $T_i = \frac{T}{K_i}$ and calculate as follows:
\begin{align}
    z^i_j &= \agg_i(\vy_\win{i,j}) = \sum_{r=1}^{K_i} a^i_r y_{r+(j-1)K_i}
\end{align}
 


Here are examples of two aggregations -- (i) average and (ii) trend. Average is just: 
\begin{align}
    \label{eqn:aggregate_sum}
    z^i_j &= \sum_{r=1}^{K_i} \underbrace{\frac{1}{K_i}}_{a^i_r} y_{(j-1)K_i + r}
\end{align}
Trend aggregate of the window captures how the series changes along time. A positive (negative) trend denotes that the value increases (decreases) as we move forward in the window.  We compute trend as the slope of a linear fit on the values in each window as 
\begin{align}
    \label{eqn:aggregate_trend} z^i_j &= \sum_{r=1}^{K_i} \underbrace{(\frac{r}{K_i} - \frac{K_i+1}{2K_i})}_{a^i_r} \cdot y_{(j-1)K_i + r}
\end{align}
 We show examples of a base-level time series and one computed aggregate (average) in Figure~\ref{fig:anecdotes}.
Our framework can work with any aggregate like the above two that can be expressed as a linear weighted sum of its arguments.  
For notational ease we refer to the base-level also as an aggregate function $A_0$ with window size $K_i=1$ and $\va^i=[1]$.

\subsection{Forecast Method}
\label{subsec:klst}
We train independent probabilistic forecasting models for the series corresponding to each aggregate $A_i \in \cA$ which includes the original time-series as well.  Our work is agnostic of the exact neural architecture used for the forecasts.  We present details of how to train these models in Section~\ref{subsec:train}.   

During inference we use the known history $H_T$ and the trained models to get forecasts $\fP(z^i_j|H_T,\vx_j)$ for each of the variables $z^i_1,\ldots,z^i_{T_i}$ for each aggregate function $A_i$, which includes the base-level quantities.

Since models for all aggregates and for the original series are trained independently, these predicted distributions are not necessarily coherent.  We infer a new consensus distribution $Q(y_{T+1}\ldots y_{T+R})$ that minimizes the KL-distance with each of the forecast distributions. We choose a tractable form for $Q$: a multivariate Gaussian with mean $\vmu=[\mu_{T+1}\ldots \mu_{T+R}]^T$ and covariance $\Sigma$.  With this form we can compute the marginal distribution of aggregate variable $z^i_j$ as:
\begin{align}
    \label{eqn:distribution_aggregate}
    Q^i_j(z^i_j|\vmu,\Sigma) = \Ncal \Big(z^i_j;\vmu_\win{i,j}^T\va^i, ~~{\va^i}^T \Sigma_\win{i,j} \va^i \Big)
\end{align}
We use the notation $\Sigma_\win{i,j}$ to denote the sub-matrix of $\Sigma$ that spans over the indices in $\win{i,j}$. Likewise for $\vmu_\win{i,j}$.

Using these we can write our objective of finding the parameters $\vmu$ and $\Sigma$ so as to minimize the KL-distance with each forecast distribution as:

\begin{equation}
    \min_{\vmu,\Sigma} \sum_{i\in\cA} \sum_{j=T_i}^{T_i+R_i} \alpha_i \KLD{Q^i_j(z^i_j|\vmu,\Sigma)}{\fP(z^i_j|\bullet)}
    \label{eqn:minKL}
\end{equation}
Here, the weights $\alpha_i$ are hyper-parameters which denote importance of KL-distance of $i$-th aggregate.

When the distribution of each forecast variable $\fP(z^i_j|\bullet)$ is represented as a Gaussian with mean $\mean(z^i_j)$ and variance $\var(z^i_j)$, their KL distance can be expressed in closed form\footnote{$\KLD{\Ncal(\mu_q,\sigma_q^2)}{\Ncal(\mu_p,\sigma_p^2)}=\frac{(\mu_q-\mu_p)^2+\sigma_q^2}{2\sigma_p^2}-\log \frac{\sigma_q}{\sigma_p}-\frac{1}{2}$}. Also, the optimization over the $\vmu$ and $\Sigma$ terms separate out into two independent optimization problems. For $\vmu$ the objective is now: 

\newcommand{\pva}{\tilde{\va}}
\begin{align}
    \min_\vmu  \sum_{i\in\cA} \sum_{j=T_i}^{T_i+R_i} \frac{1}{\var(z^i_j)^2}(\vmu_\win{i,j}^T\va^i-\mean(z^i_j))^2
\end{align}
This is a convex quadratic objective that can be solved in closed form as
\begin{align}
\label{eqn:mean_solve}
\vmu^* = \big[ \sum_{i\in\cA} \sum_{j=T_i}^{T_i+R_i} \frac{\pva_j^i{\pva_j^i}{}^T}{\var(z^i_j)^2}\big]^{-1} \big[ \sum_{i\in\cA} \sum_{j=T_i}^{T_i+R_i} {\pva^i_j}{}^T\hat{\vmu}  \big]
\end{align}
where we use $\pva^i_j \in \real^{R}$ to denote the padded version of the indices $\win{i,j}$.

For $\Sigma$ the objective is:
\begin{align}
\label{eq:sigmaObj}
    \min_\Sigma  \sum_{i\in\cA} \sum_{j=T_i}^{T_i+R_i} \frac{ {\va^i}^T \Sigma_\win{i,j} \va^i }{2\var(z^i_j)^2} - \log ({\va^i}^T \Sigma_\win{i,j} \va^i)  
\end{align}
This objective cannot be solved in closed form but we follow a strategy to efficiently approximate it.  
We consider a simplification of $\Sigma$ to be a low-rank matrix. Following \cite{salinas2019high} we express the covariance $\Sigma$ in a tractable format as follows:

\begin{align*}
    \bm{\Sigma} &= \mathrm{Diag}(\sigma^2) + VV^T \\
    &=
\begin{pmatrix}
    \sigma^2_{T+1} & \ldots & 0 \\
      & \ddots &  \\
    0 & \ldots & \sigma^2_{T+R}
\end{pmatrix}
+
\begin{pmatrix}
    v_{T+1} \\
    \vdots  \\
    v_{T+R}
\end{pmatrix}
\begin{pmatrix}
    v_{T+1} \\
    \vdots  \\
    v_{T+R}
\end{pmatrix}^T
\end{align*}
where $v_{T+r} \in \real^k$ is $r$-th row of $V$, and $k$ is the chosen rank of the covariance matrix.

Using the above form of $\Sigma$, Equation~\ref{eq:sigmaObj} can be rewritten as:
\begin{align}
    \label{eqn:sigmaObjbreakdown}
    \min_{V,\sigma}
    \sum_{i\in\cA} \sum_{j=T_i}^{T_i+R_i} \frac{ {\pva_j^i}{}^T[\text{Diag}(\sigma^2)] \pva_j^i + ({\pva_j^i}{}^T V)^2 }{2\var(z^i_j)^2} \\ - \log \big({\pva_j^i}{}^T[\text{Diag}(\sigma^2)] \pva_j^i + ({\pva_j^i}{}^T V)^2\big)  
\end{align}

With the above representation of the joint distribution $Q(y_T,\ldots y_{T+R})$, the number of parameters needed to store the distribution is linear in $R$, and also we can compute the distribution of any linear aggregate efficiently. Algorithm~\ref{alg:klst_inference} presents an overview of our method.

\begin{algorithm}[t] 
	\caption{\Our\ Training and Inference Algorithm}
	\label{alg:klst_inference}
	\begin{algorithmic}[1]
	\STATE \textbf{Input: }  Training data $\{(\vx_1,y_1)\ldots (\vx_n,\vy_n)\}$, aggregate functions $\cA=\{(\va_i,K_i)\}$, forecast horizon $R$. 
    \FOR{$A_i$ in $\cA$}
        \STATE \texttt{/*Train $i$-th aggregate model using training data for $i$-th aggregate */}
        \STATE $\Dtrn^i = z^i_1,\ldots,z^i_{n/K_i}$.
        \STATE $\theta^i \leftarrow \textsc{Train}(\Dtrn^i)$ using Eqn~\ref{eqn:train}
        \STATE Get forecasts $\fP(z_j^i | \theta^i)$ for $j=T_i+1,\ldots,T_i+R_i$.
    \ENDFOR
    \STATE Get coherent mean forecasts $\vmu^*$ using Eqn.~\ref{eqn:mean_solve}.
    \STATE Get coherent variance forecasts ($V^*, \sigma^*$) using Eqn.~\ref{eqn:sigmaObjbreakdown}.
    \STATE \texttt{/* End of Training */}
    \STATE Inference: Given new aggregate $A_{\mathrm{new}}$ defined via $\tilde{\va}^{\mathrm{new}}$ 
    \STATE ${\vmu^*}^{\mathrm{new}} = {\vmu^*}{}^T \tilde{\va}^{\mathrm{new}}$.
    \STATE $\Sigma^*{}^{\mathrm{new}} = \sum_{j=1}^R ({\tilde{a}^{\mathrm{new}}_j})^2 \sigma_j^{*}{}^{2} + ({\tilde{\va}^{\mathrm{new}}})^T V^*V^*{}^T ({\tilde{\va}^{\mathrm{new}}}) $
	\end{algorithmic}
\end{algorithm}

\subsection{Training Parameters of Forecast Models}
\label{subsec:train}
As mentioned earlier we train separate models for each aggregate series.   For each aggregate $\agg_i$ the forecast model can be viewed as a function $F(H_T, \vx_R|\theta^i)$ of the history of known values $H_T=(\vx_1,y_1)\ldots,(\vx_T,y_T)$, and known input features $\vx_{T+1}\ldots \vx_{T+R}$ of the next $R$ values for which the model needs to forecast.  We use $\theta^i$ to denote the parameters of the $i$th forecast model.  The exact form of the forecast model $F()$ that we used is described next, followed by a description of the training procedure.

\subsubsection{Architecture of Forecast Model}
\label{subsec:architecture}
Our forecasting model $F(H_T, \vx_R|\theta^i)$ is based on Transformers like Informer~\cite{zhou2020informer} but with subtle differences that provide gains even for the baseline predictions.

Our model first applies a convolution on the input $\vz$ series and features $\vx$. The Transformer's multi-headed attention layers are on a concatenation of the convolution layer output.
\begin{align*}
    \vz^c &= \mathrm{Conv}(\vz, w_y, 1) \quad \vx^c = \mathrm{Conv}(\vx, w_f, 1) \\
    \vh &= \mathrm{Transformer}([\mathrm{Concat}(\vz^c_t, \vx^c_t) + \ve_t:t=1\ldots T])
\end{align*}
where $\vh \in \RR^{T \times d_{\mathrm{model}}}$ is the output of the transformer encoder and $\ve_t$ denotes the positional encoding of time index $t$. 
In the decoder, we \emph{warm start} to provide more context to the decoder apart from the encoder state $\vh$. In warm start, we select the last $s$ values in the input series and use them in the decoder of the transformer.
\begin{align*}
    \vz^c &= \mathrm{CONV}([\vz_{T-s+1\ldots T}, \mathbf{0}_R], w_z, 1) \\
    \vx^c &= \mathrm{CONV}([\vx_{T-s+1\ldots T}, \vx_{T+1\ldots T+R}], w_x, 1) \\
    \vh_{\mu} &= \mathrm{TransformerDecoder}_{\mu}(\mathrm{Concat}(\vz^c, \vx^c) + \ve, \vh) \\
    \vh_{\sigma} &= \mathrm{TransformerDecoder}_{\sigma}(\mathrm{Concat}(\vz^c, \vx^c) + \ve, \vh) \\
    \mu &= \mathrm{Linear}(\vh_{\mu}) \quad \sigma = \mathrm{Softplus}(\mathrm{Linear}(\vh_{\sigma}))
\end{align*}
where $\mathbf{0}_R$ is a vector of zeros of length $R$, $\vh_{\mu} \in \RR^{R \times d_{\mathrm{model}}}$, $\vh_{\sigma} \in \RR^{R \times d_{\mathrm{model}}}$, $\mu \in \RR^{R \times 1}$, and $\sigma \in \RR^{R \times 1}.$

\subsubsection{Training Procedure}
We start with base level series $(\vx_1, y_1)\ldots (\vx_n,y_n)$. First we split the series into training, validation, and test sets of lengths $\trnlen$, $\vallen$, and $\testlen$ respectively as follows:
\begin{align*}
    \Dtrn &= \{(\vx_t, y_t)| t=1,\ldots, \trnlen \} \\
    \Dval &= \{(\vx_t, y_t)| t=\trnlen-T+1,\ldots,\trnlen+\vallen \} \\
    \Dtest &= \{(\vx_t, y_t)| t=\trnlen+\vallen-T+1, \ldots, n \}
\end{align*} 
Note that when $\vallen \ge R$ or $\testlen \ge R$, we use rolling-window setting as done in \cite{Deshpande2019,Rangapuram2018} on validation and test sets.

We create training batches by sampling \emph{chunks} of length $T+R$ from training set $\Dtrn$ as 
\begin{align*}
\{(\vx_t, y_t)&| t=p,\ldots p+T+R \} \\
&~~\text{where}~~p \sim [1,\ldots, \trnlen-(T+R-1)]
\end{align*}

Similarly, we process $i$-th aggregate series $z^i_1,\ldots, z^i_{n/K_i}$. Validation and test sets across base-level and aggregated series are aligned. To aggregate features, we simply take average value of each feature in the aggregation window for all aggregates. For an aggregate model for $i$-th aggregate with window size $K_i$, the forecast horizon is $R_i=R/K_i$. Similarly, the size of the history also changes. We set $T_i=B(T/K_i)$, so the aggregate model can effectively see the history $B$-times the size of base level model's history. We found setting $B$=2 works well in practice. It is also a good option for smaller datasets such as \etth. 

For each aggregate function $\agg_i$, we sample training chunks of size $T_i+R_i$ and train the parameters $\theta^i$ of the Transformer model for the following training objective.
\begin{align}
\label{eqn:train}
    \max_{\theta^i} \sum_{(\vx^i_j,\vz^i_j)} \sum_{t=T_i+1}^{T_i+R_i} \log \Ncal(z_t ; (\mu_t, \sigma_t)=F(H_T,\vx,t|\theta^i))
\end{align}

\begin{figure*}
    \centering
    \includegraphics[width=0.49\textwidth]{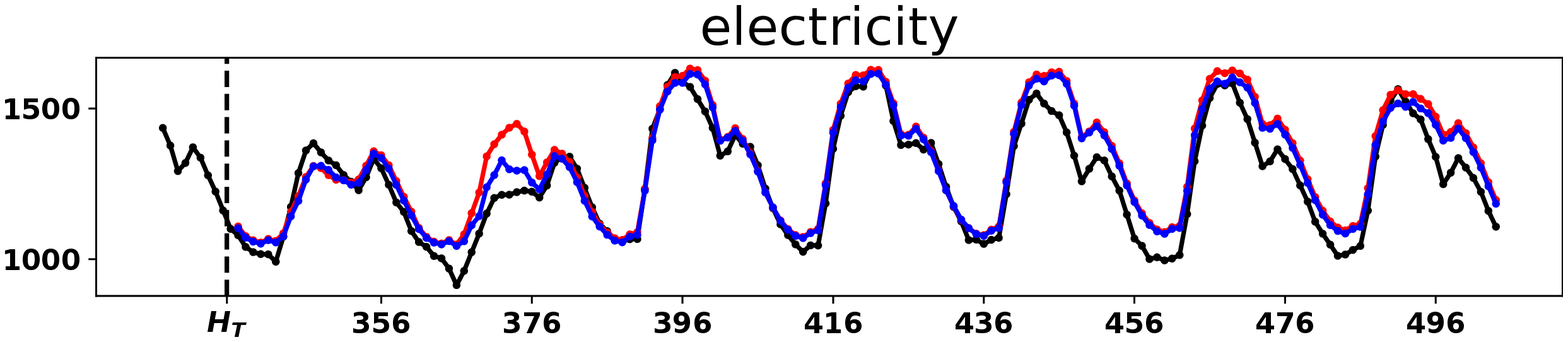}
    \includegraphics[width=0.49\textwidth,height=0.11\textwidth]{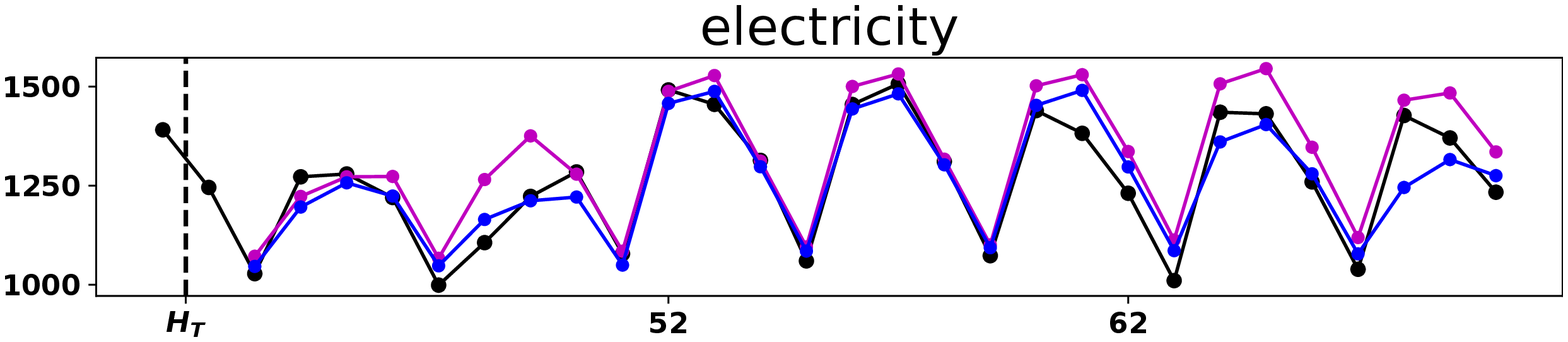}
    \includegraphics[width=0.49\textwidth,height=0.11\textwidth]{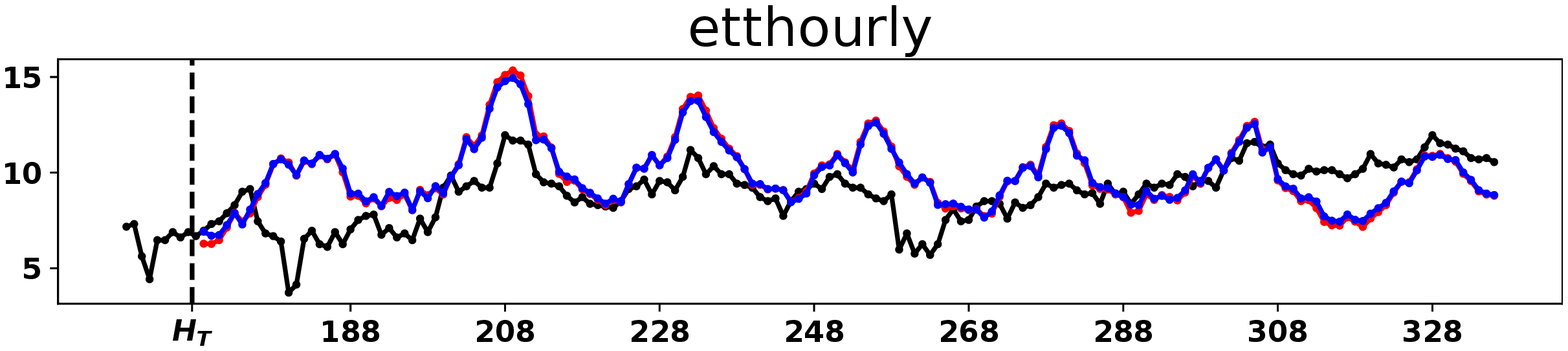}
    \includegraphics[width=0.49\textwidth,height=0.11\textwidth]{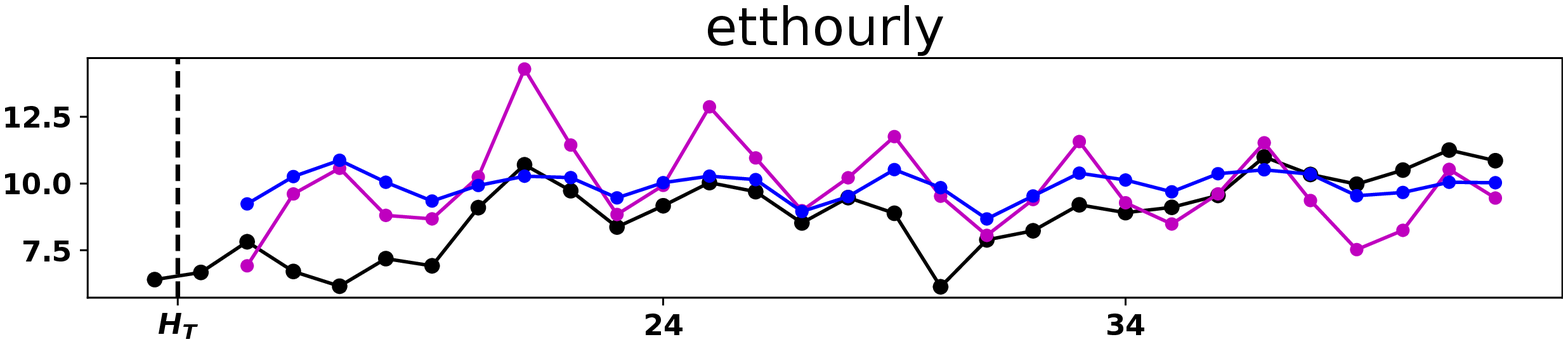}
    \includegraphics[width=0.49\textwidth,height=0.11\textwidth]{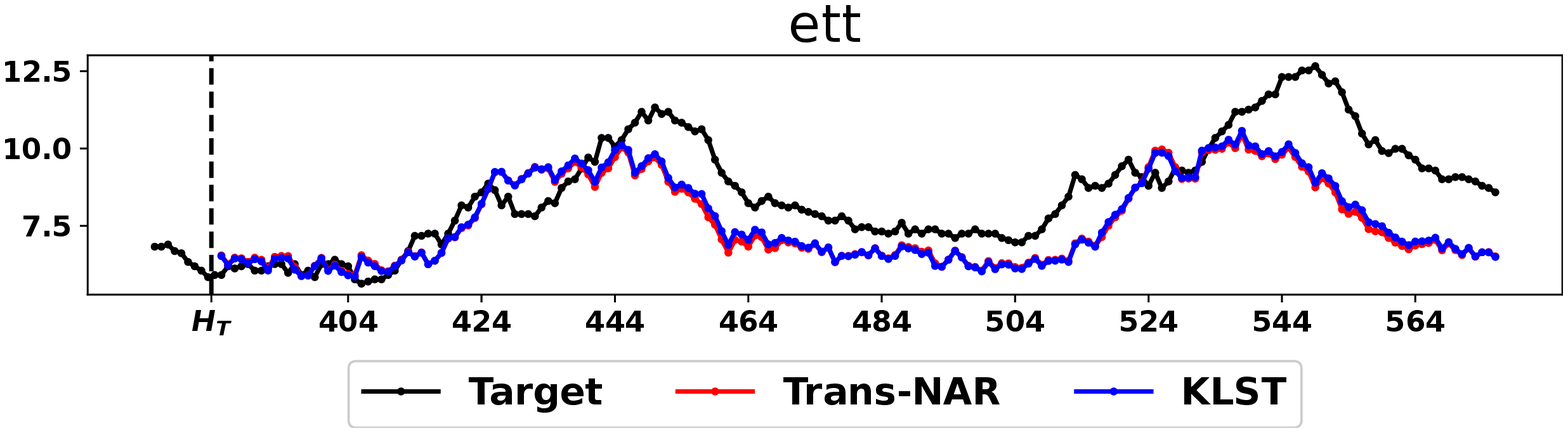}
    \includegraphics[width=0.49\textwidth,height=0.11\textwidth]{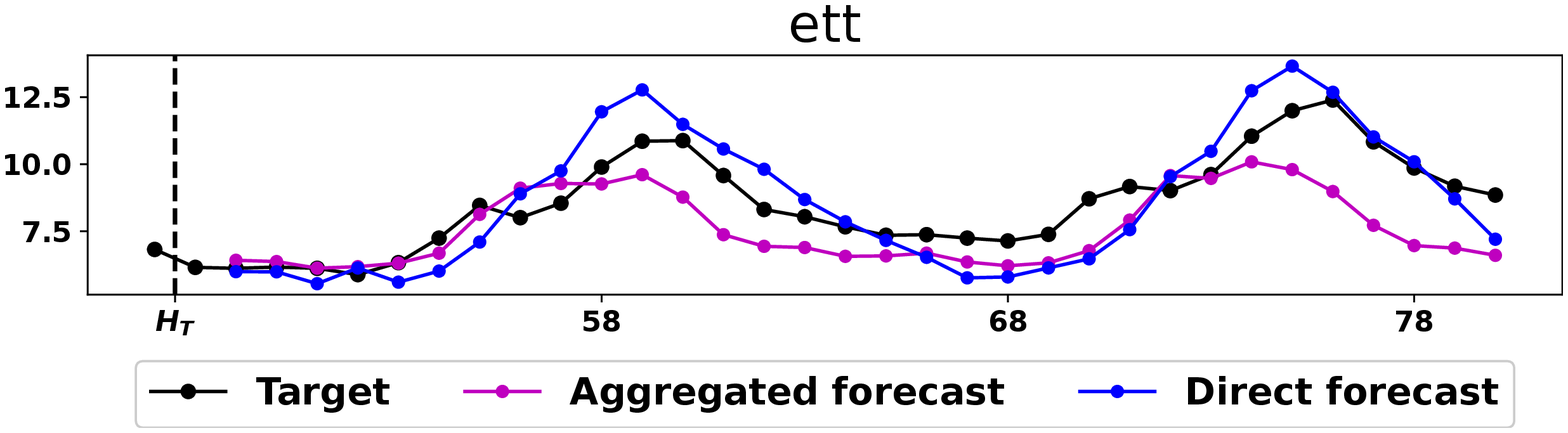}
    \caption{Left Column: Base level forecasts, Right Column: Forecasts of \sumagg\ aggregate with $K=6$.}
    \label{fig:anecdotes}
\end{figure*}
\vspace{-4mm}
\section{Related Work}
Time series forecasting is an extensively researched problem. In deep learning, the problem got renewed interest~\cite{benidis2020neural} after an RNN-based auto-regressive model~\cite{FlunkertSG17,Mukherjee2018} established conclusive gains over conventional machine learning and statistical methods.  While these early neural models were based on RNNs, more recently Transformer-based models have been found to provide faster and efficient predictions on time-series data~\cite{li2019enhancing,lim2020temporal,bansal2021missing,zhou2020informer}. For long-term forecasting auto-regressive models  are both slow and subject to cascading errors from previous predicted values for next-step predictions.  Another option is non-auto-regressive models (NAR) which predict each future forecast independently in parallel~\cite{wen2017multi,Deshpande2019,zhou2020informer}. However, since NAR models do not capture the joint distribution, they fail to provide accurate results for aggregate queries on forecasts.  N-beats~\cite{oreshkin2019n} is another recent architecture for time-series forecasting based on layered residual connections, specifically designed for interpretability.  Our technique is orthogonal to the underlying time-series forecasting model.  \nocite{gasthaus2019probabilistic}

\paragraph{Multi-level Models}
Our approach of independently predicting models at different aggregation levels and then reconciling is related to the approach in \cite{taieb2017coherent,taieb2019,Wickramasuriya2019}.  Their hierarchies are over items and consider only sum-based aggregates whereas we create hierarchies along time and consider arbitrary linear aggregates. Also, our objective is to minimize the KL distance with the distribution at various levels whereas they focus on error minimization and  their parameterization of the reconciliation is different from ours.  \cite{han21a} is another multi-level model for item hierarchies that train models in a bottom up manner with higher level models regularized to be consistent with lower layers. They do not model inter-item correlation, can only efficiently handle fixed aggregates, and do not revise base-level forecasts.  We obtain gains at the base-level too via the hierarchies as we show in our empirical comparison. \cite{rangapuram2021end} propose an end-to-end method that produces coherent forecasts. However, their method relies on the sampling to obtain covariance of the joint distribution whereas we train the covariance through our inference method alone.
%
%

\paragraph{Beyond Base-level Forecasting Loss}
Our method of learning the slope as an aggregate in the forecast horizon is related to recent efforts at trying to learn the shape of the series. Recently, \cite{Guen2019,guen2020probabilistic} learn the shape of the series in forecast horizon using a differentiable approximation of Dynamic Time Warping (DTW) loss function. They also contain additional losses that minimize distortion and increase diversity. 
SRU captures moving statistics of the history through a recurrent layer \cite{OlivaPS17}. However, they use the statistics only to capture dependence among observed variables.

\paragraph{Long-term Forecasting Model}
Recently \cite{zhou2020informer} proposed a Transformer architecture for long-range time-series forecasting. Their focus is to increase efficiency of the self-attention in Transformers for long series by enforcing sparsity. 
To the best of our knowledge, this is the best performing of existing methods for long-term forecasting.  We compare empirically with this model and present significant gains on multiple datasets.   
\definecolor{Gray}{gray}{0.9}
\definecolor{LightCyan}{rgb}{1,1,0.6}
\newcolumntype{g}{>{\columncolor{Gray}}c}
\newcolumntype{y}{>{\columncolor{LightCyan}}c}
\begin{table}[t!]
\small
\centering
\tabcolsep=0.13cm
\begin{tabular}{|l|l|y|r|r|r|r|}
\hline
Dataset & \multirow{2}{*}{Model} & \multicolumn{5}{c|}{K} \\ \cline{3-7}
Agg & & 1 & 4 & 8 & 12 & 24 \\ \hline

\ett & Informer & 7.01 & 7.00 & 7.00 & 7.00 & 6.98 \\
\sumagg & \TransformerAR & 3.03 & 3.29 & 3.38 & 3.38 & 3.43 \\
& \TransformerNAR & 1.25 & 1.36 & 1.39 & 1.39 & 1.38 \\
& \sharq & 1.25 & 1.87 & 1.78 & 1.80 & 1.82 \\
& \Our & \textbf{1.17} & \textbf{1.14} & \textbf{1.17} &\textbf{ 1.19} & \textbf{1.22} \\
\hline

\ett & \TransformerNAR & 1.25 & \textbf{0.13} & \textbf{0.07} & \textbf{0.06} & 0.05 \\
\slopeagg & \Our & \textbf{1.17} & 0.30 & 0.12 & \textbf{0.06} & \textbf{0.04} \\
\hline

\ett & \TransformerNAR & 1.25 & \textbf{0.14} & \textbf{0.16} & \textbf{0.20} & 0.29 \\
\diffagg & \Our & \textbf{1.17} & 0.33 & 0.26 & 0.25 & \textbf{0.26} \\
\hline \hline

\solar & Informer & 41.02 & 36.31 & 34.85 & 17.55 & 13.14 \\
\sumagg & \TransformerAR & 21.13 & 18.91 & 18.40 & 16.37 & 16.17 \\
& \TransformerNAR & 13.85 & 13.25 & 12.95 & 12.78 & 12.43 \\
& \sharq & 13.85 & 13.36 & 13.22 & 14.21 & \textbf{11.60} \\
& \Our & \textbf{12.95} & \textbf{12.73} & \textbf{12.54} & \textbf{12.43} & 12.21 \\
\hline

\solar & \TransformerNAR & 13.85 & 4.86 & 3.02 & 4.10 & 0.39 \\
\slopeagg & \Our & \textbf{12.95} & \textbf{4.49} & \textbf{2.82} & \textbf{3.98} & \textbf{0.37} \\
\hline

\solar & \TransformerNAR & 13.85 & 5.03 & 5.98 & 12.59 & 5.62 \\
\diffagg & \Our & \textbf{12.95} & \textbf{4.63} & \textbf{5.53} & \textbf{12.23} & \textbf{5.35} \\
\hline \hline

\etth & Informer & 4.80 & 4.77 & 4.73 & 4.67 & 4.57 \\
\sumagg & \TransformerAR & 1.96 & 2.01 & 1.98 & 2.01 & 1.96 \\
& \TransformerNAR & 1.79 & 1.92 & 1.93 & 1.92 & 1.89 \\
& \sharq & 1.79 & 1.91 & 1.73 & 1.75 & 1.78 \\
& \Our & \textbf{1.64} & \textbf{1.61} & \textbf{1.65} & \textbf{1.67} & \textbf{1.69} \\
\hline 

\etth & \TransformerNAR & 1.79 & \textbf{0.26} & 0.20 & 0.14 & 0.07 \\
\slopeagg & \Our & \textbf{1.64} & 0.37 & \textbf{0.18} & \textbf{0.11} & \textbf{0.06} \\
\hline

\etth & \TransformerNAR & 1.79 & \textbf{0.27} & 0.39 & 0.46 & 0.50 \\
\diffagg & \Our & \textbf{1.64} & 0.40 & \textbf{0.37} & \textbf{0.39} & \textbf{0.41} \\
\hline \hline

\electricityshort & Informer & 172.3 & 159.7 & 155.8 & 118.1 & 109.6 \\
\sumagg & \TransformerAR & 140.2 & 137.8 & 134.0 & 109.6 & 104.7 \\
& \TransformerNAR & 54.1 & 53.5 & 52.3 & 50.8 & 48.4 \\
& \sharq & 54.1 & \textbf{49.8} & \textbf{47.0} & 50.5 & 46.3 \\
& \Our & \textbf{50.2} & 50.6 & 49.6 & \textbf{48.4} & \textbf{46.2} \\
\hline

\electricityshort & \TransformerNAR & 54.1 & 8.96 & 6.25 & 5.65 & 2.23 \\
\slopeagg & \Our & \textbf{50.2} & \textbf{8.26} & \textbf{5.76} & \textbf{5.18} & \textbf{2.14} \\
\hline

\electricityshort & \TransformerNAR & 54.1 & 9.50 & 13.23 & 18.37 & 16.13 \\
\diffagg & \Our & \textbf{50.2} & \textbf{8.80} & \textbf{12.22} & \textbf{16.84} & \textbf{15.44} \\
\hline

\end{tabular}
\caption{Comparison of \our\ using CRPS -- (i) with all baselines on Long-Term Forecasts (Column with $K$=1). (ii) with \TransformerNAR\ on unseen aggregates (Columns $K$=4,8,12,24).}
\label{tab:main_table_merged}
\end{table}

\section{Experiments}
We evaluate our method on four real-life datasets and contrast with state-of-the-art methods of long-term forecasting.

\subsection{Datasets}

\paragraph{Electricity} 
This dataset contains national electricity load at Panama power system\footnote{\url{https://data.mendeley.com/datasets/byx7sztj59/1}}. The dataset is collected at an hourly granularity from January 2015 to March 2020. 

\paragraph{Solar}
This dataset is on hourly photo-voltaic production of $137$ stations and was used in \cite{salinas2019high}.

\paragraph{ETT (Hourly and 15-minute)} 
This dataset contain a time-series of oil-temperature at an electrical transformer \footnote{\url{https://github.com/zhouhaoyi/ETDataset}} collected from July 2016 to June 2017. The dataset is available at two granularities --- $15$ minutes (\ett) and one hour (\etth). Both were used   in \cite{zhou2020informer}.

\begin{table}[b!]
\small{
    \centering
    \tabcolsep=0.13cm
    \begin{tabular}{|l|r|r|r|r|r|}
    \hline
    \multirow{2}{*}{Dataset} & \# & Avg. & \multirow{2}{*}{$R$} & train-len. & test-len. \\
      & Series &  $T$ & & /series & /series \\ \hline
    \ett & 1 & 384  & 192 & 55776 & 13824 \\
    \etth & 1 & 168 & 168 & 14040 & 3360 \\
    \electricity & 1 & 336 & 168 & 36624 & 9072 \\
    \solar & 137 & 336 & 168 & 7009 & 168 \\ \hline
    \end{tabular}
    \caption{Summary of the datasets}
    \label{tab:dataset_summary}
}
\end{table}

\subsection{Base Models}
\label{subsec:base_models}
\paragraph{Informer}~\cite{zhou2020informer}
is a recent Transformer-based model specifically designed for long-term forecasting. They also use a non-auto-regressive model and predict each future value independently.

\paragraph{\TransformerNAR}
refers to our Transformer architecture described in Section~\ref{subsec:architecture} but where we predict only a single model on only the base-level forecasts.  

\paragraph{\TransformerAR}
\TransformerAR\ is an autoregressive version of the transformer architecture described in Section~\ref{subsec:architecture}. 
At the $t$-th step during prediction, it performs convolution on the window $\hat{z}_{t-w_y+1},\ldots,\hat{z}_{t-1}$ of predicted values. When $t<w_f$, the convolution window overlaps with the history $H_T$ and observed values from $H_T$ is used. 

\paragraph{\sharq} SHARQ \cite{han21a} is a hierarchical forecasting method in which a regularization loss ensures that the coherency error between aggregate predictions and aggregation of base-level predictions is minimized. Since \sharq\ is agnostic to the underlying forecasting model, we use our baseline \TransformerNAR\ as the base model for \sharq. For $j$-th value of an aggregate model, we write the regularization term of \sharq\ as:
\begin{align}
\label{eqn:sharq_loss}
\lambda_i(\hat{\mu}-\va^T \vmu_{w_{\cdot,j}})^2 + \big((\hat{q}_{\tau}-\hat{\mu})^2 - \va^T(\hat{q}_{\tau}-\vmu_{w_{\cdot,j}}\big)^2\va)^2
\end{align}
where $\hat{\mu}$ and $\hat{q}_{\tau}$ denote the predicted mean and aggregate model at $j$-th position. This model requires a fresh model to be trained for each aggregate unlike our method.

\paragraph{\sysname} is our method as described in Algorithm~\ref{alg:klst_inference}.
The choice of aggregate functions and range of $K$ values are based on validation set and chosen from aggregate functions = \{ \sumagg, \slopeagg \} and  $K$=\{6,12\}.
For all datasets, we assign weights $\alpha$=\{10,0.5\} to \sumagg\ and \slopeagg\ aggregates respectively. Since direct predictions on \sumagg\ aggregate are often more accurate, we assign higher weight to it.

\vspace{-1mm}
\subsection{Evaluation Metric}

We evaluate probabilistic forecasts using the well-known~\cite{salinas2019high} Continuous Ranked Probability Score (CRPS) metric. 
If we denote the predicted CDF for a target value $y_t$ by $F_t(\cdot)$, we can write the CRPS by integrating the quantile loss over all possible quantiles~\cite{salinas2019high}:
\begin{align*}
    \mathrm{CRPS}(F_t^{-1}, y) &= \int_0^1 2 \big(\alpha-\mathcal{I}_{[F_t^{-1}(\alpha)<q]}\big)(y_t-q) d\alpha
\end{align*}
It reduces to Mean Absolute Error (MAE) when the forecast is a point,
and is a proper scoring metric of the quality of an entire distribution including its mean and variance.

\subsection{Qualitative Results}

We present some anecdotes to provide insights on why \our\ should enhance the accuracy of base-level forecasts.  In Figure~\ref{fig:anecdotes} (left column) we show that forecasts by \our\ (blue) are closer to the ground truth (black) compared to the base-level model \TransformerNAR\ (red).  We explain these gains by showing the improved accuracy of the model that independently predicts the sum aggregate for a window of $K=6$  in the right column of Figure~\ref{fig:anecdotes}. 
We observe that direct forecast values are much closer to the ground-truth than those from aggregating base-level forecasts.  When these direct aggregates are used to revise the base-level forecasts to be consistent with these aggregates (as described in Section~\ref{subsec:klst}), we obtain better forecasts at the base-level. 

\subsection{Comparison of \our\ with Baselines}
\paragraph{Impact of \Our\ on Long-Term Forecasts}
In Table~\ref{tab:main_table_merged}, we compare \our\ with the baselines described in Sec.~\ref{subsec:base_models}. Under \sumagg\ aggregate and column $K$=1 (highlighted), we compare the CRPS of base level forecasts for Informer, \TransformerAR, \TransformerNAR, \sharq\ and \our. For base-level forecasts, \our\ performs better than all baselines while \TransformerNAR\ and \sharq\ are share the second position. Since \sharq\ does not update base level model parameters, base level forecasts of \sharq\ are identical to \TransformerNAR.

\paragraph{Results on Unseen Aggregates Post Inference}
An important aspect of time-series analytics is the effectiveness of base-level forecasts on various types of aggregates at various granularities. Training an aggregate model for each aggregate and at each granularity is rather expensive. Our approach is to train a few representative aggregate models using a fixed set of $K_i$ values. Then, forecasts after consensus can be used to compute other aggregate and at other granularities. 

In Table~\ref{tab:main_table_merged} (columns with $K$=4,8,12,24), we compare \Our\ with (i) all baselines for \sumagg\ aggregate and (ii) with \TransformerNAR\ for \slopeagg\ and \diffagg\ aggregates. The set of $K$ values we use in Table~\ref{tab:main_table_merged}
also contain the $K$ for which no aggregate model is trained. Also, no aggregate model was trained corresponding to the \diffagg\ aggregate. For each aggregate and for each $K$, first we aggregate the forecasts of baselines and \our\ and then evaluate the aggregates using CRPS. Since \sharq\ does not update the bottom level forecasts, we have trained a separate aggregate model for each $K$ and \sumagg\ aggregate using \sharq\ regularization loss (Eqn.~\ref{eqn:sharq_loss}). 

For \sumagg\ aggregate, \our\ outperforms all baselines on \ett\ and \etth\ dataset. \sharq\ performs better than \our\ on \solar\ and \electricity\ for $K$=24 and $K$=4,8 respectively. Since \sharq\ requires training a separate model for each aggregate and $K$, we compare \sharq\ only on \sumagg\ aggregate.

For \slopeagg\ and \diffagg\ aggregates, \our\ outperforms \TransformerNAR\ on \solar\ and \electricity. Whereas for \ett\ and \etth, \our\ is better than \TransformerNAR\ in 12 out of 20 cases. 

\section{Conclusion}
In this paper we addressed the problem of long-range forecasting with the goal of providing accurate probabilistic predictions, not just at the base-level but also at dynamically chosen aggregates.  We proposed a simple method of independently training forecast models at different aggregation levels, and then designing a consensus distribution to minimize distance with forecasts from each component distribution.  We obtained significant gains in accuracy over the base-level and over new aggregate functions at new granularity levels defined during test time. Compared to existing hierarchical methods, we impact both base-level predictions and new aggregates defined dynamically.  

\bibliographystyle{named}
\bibliography{ML,pubs,refs}

\end{document}